\pdfoutput=1

\documentclass[11pt]{article}

\usepackage[]{EMNLP2022}

\usepackage{times}
\usepackage{latexsym}

\usepackage[T1]{fontenc}

\usepackage[utf8]{inputenc}

\usepackage{microtype}

\usepackage{inconsolata}
\usepackage{upgreek}
\usepackage{color}
\usepackage{colortbl}
\usepackage{amsmath}
\usepackage{amssymb}
\usepackage{hyperref}
\usepackage{algorithm}
\usepackage{algorithmic}
\usepackage{enumitem}
\usepackage{booktabs}
\usepackage{multicol}
\usepackage{multirow}
\usepackage{float}
\usepackage{graphics}
\usepackage{graphicx}
\usepackage{makecell}
\usepackage{array}
\usepackage{bm}
\usepackage{booktabs}
\usepackage{ bbold }
\usepackage{bbding}
\newcommand{\secref}[1]{\S\ref{#1}}

\title{Semi-Supervised Lifelong Language Learning}

\author{\textbf{Yingxiu Zhao$^{1}$\thanks{\quad Work done while the author was interning at Alibaba.}, \ 
Yinhe Zheng$^2$\thanks{\quad Corresponding author.}, \ 
Bowen Yu$^2$, \ 
Zhiliang Tian$^1$}, \\
\textbf{
Dongkyu Lee$^1$, \
Jian Sun$^2$, \ 
Haiyang Yu$^2$,
Yongbin Li$^2$\footnotemark[2], \ 
Nevin L. Zhang$^1$} \\
$^1$ The Hong Kong University of Science and Technology, Hong Kong, \\
$^2$ Alibaba Group, China \\
\small \texttt{\{yzhaocx,ztianac,dleear,lzhang\}@connect.ust.hk},
\texttt{zhengyinhe1@163.com},\\
\small \texttt{\{yubowen.ybw,yifei.yhy,shuide.lyb\}@alibaba-inc.com, jiansun\_china@hotmail.com}
}

\begin{document}
\maketitle
\begin{abstract}
Lifelong learning aims to accumulate knowledge and alleviate catastrophic forgetting when learning tasks sequentially. However, existing lifelong language learning methods only focus on the supervised learning setting. Unlabeled data, which can be easily accessed in real-world scenarios, are underexplored. In this paper, we explore a novel setting, semi-supervised lifelong language learning (SSLL), where a model learns sequentially arriving language tasks with both labeled and unlabeled data. We propose an unlabeled data enhanced lifelong learner to explore SSLL. Specially, we dedicate task-specific modules to alleviate catastrophic forgetting and design two modules to exploit unlabeled data: (1) a virtual supervision enhanced task solver is constructed on a teacher-student framework to mine the underlying knowledge from unlabeled data; and (2) a backward augmented learner is built to encourage knowledge transfer from newly arrived unlabeled data to previous tasks. Experimental results on various language tasks demonstrate our model's effectiveness and superiority over competitive baselines under the new setting SSLL.
We will release the code and data \footnote{\url{https://github.com/AlibabaResearch/DAMO-ConvAI/tree/main/ssll}}.

\end{abstract}

\section{Introduction}
\label{sec:intro}
A remarkable ability of humans is to learn and accumulate knowledge continuously throughout their lifetime. 
Such \textit{Lifelong Learning} ability is crucial for computational systems interacting with the real world and processing continuous streams of information~\cite{parisi2019continual,defy}.
However, most deep neural networks studies assume data distributions are stationary, which is not applicable in the real-world environments that dynamically evolve.
In such real scenarios, models often suffer from \textit{catastrophic forgetting} \citep{forget2,parisi2019continual}: 
a phenomenon where models forget previously learned knowledge when learning new tasks sequentially.

Various approaches have been proposed to alleviate catastrophic forgetting in lifelong scenarios. Attempts include constraining the variants of important weights with regularization~\cite{ewc,mifei}, storing real samples or using pseudo samples for previous tasks to maintain the learned knowledge~\cite{episodic,lamol,l2kd}, or dedicating task-specific modules to avoid the interference among tasks~\citep{tod,lfpt5,adapter_composi}.
Despite their reported effectiveness, these approaches are mostly designed to handle supervised learning tasks, where only labeled data are available.
In real-world scenarios, labeled data are generally expensive and time-consuming to obtain, whereas unlabeled data are much easier to collect. These unlabeled data often carry rich information and have been successfully utilized to improve model performance in semi-supervised learning \citep{uda,chen2021revisiting}.
\begin{figure}[t]
    \centering
    \includegraphics[width=0.45\textwidth]{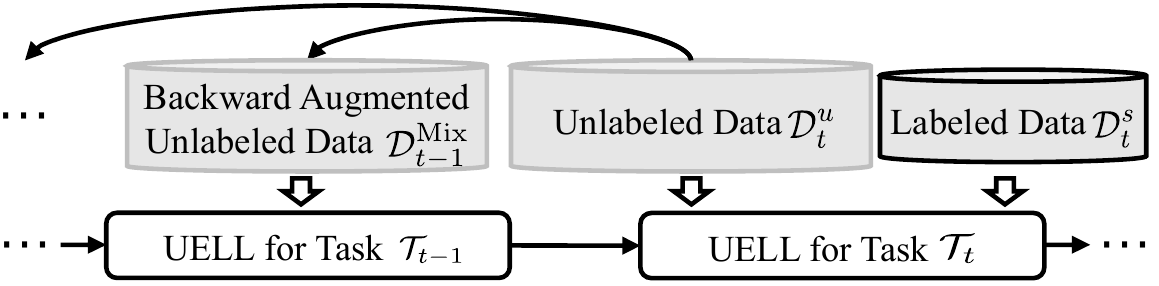}
    \caption{The Training process of our model UELL.}
    \label{fig:data_stream}
\end{figure}

In this paper, we investigate a novel setting: \underline{S}emi-\underline{S}upervised \underline{L}ifelong \underline{L}anguage learning (SSLL), where a model learns sequentially arriving language tasks with limited labeled data and adequate unlabeled data (see the training process in Fig.~\ref{fig:data_stream}).
The abundant information in the unlabeled data can not only facilitate learning the current task but also benefit learned tasks with similar data distributions.
For example, sentiment analysis and topic classification tasks may only differ in their label spaces while sharing the same set of unlabeled data. We can transfer knowledge among the two kinds of tasks using these unlabeled data \cite{liu2019multi}.
This phenomenon naturally leads to two challenges to be faced in the SSLL scenario:
\textbf{(1)} \textbf{\textit{{How to fully exploit unlabeled data to facilitate each arrived language task?}}} and 
\textbf{(2)} \textbf{\textit{{How to leverage newly arrived unlabeled data to encourage knowledge transfer to previous tasks?}}}

With this in mind, we propose an \underline{U}nlabeled data \underline{E}nhanced \underline{L}ifelong \underline{L}earner (UELL) framework to explore the new setting SSLL.
Specifically, we dedicate task-specific parameters to alleviate catastrophic forgetting in UELL. 
We construct two modules to tackle the challenges mentioned above.
The first module is a virtual supervision enhanced solver that exploits unlabeled data using a teacher-student framework.
The teacher generates pseudo labels for unlabeled data as virtual supervision and guides the student according to its learning progress. The student also learns from pseudo labels through self-study.
The second module is a backward augmented learner that encourages knowledge transfer from the current task to previously learned tasks.
The generated pseudo data for each learned task are augmented by retrieving semantically similar unlabeled samples from the current task, where the latter are leveraged to transfer knowledge backward to previously learned tasks.
We conduct extensive experiments and analyses on both language understanding and generation tasks and demonstrate that UELL can effectively address the challenges of SSLL.

Our main contributions are as follows:
\begin{itemize}[leftmargin=*]
\setlength{\itemsep}{0pt}
\setlength{\parsep}{0pt}
\setlength{\parskip}{0pt}
\item To the best of our knowledge, we are the first to explore the semi-supervised lifelong language learning setting, where a model learns sequentially arriving language tasks with a mixture of labeled and unlabeled data.
\item We propose a novel method, Unlabeled data Enhanced Lifelong Learner, to exploit unlabeled data and encourage backward knowledge transfer in SSLL.
\item We conduct adequate experiments and analyses on various language tasks. The results demonstrate the effectiveness of UELL over competitive baselines adapted from lifelong learning methods to the SSLL setting.
\item We believe our new paradigm imposes new challenges and opens up new research opportunities for the NLP community.
\end{itemize}

\section{Related Work}
\paragraph{Semi-Supervised Learning} 
aims to learn from both labeled and unlabeled data \citep{chapelle2009semi,semisurvey,he2022space}. As an efficient approach for semi-supervised learning, pseudo labeling \citep{zhou2010semi,pseudolabel} tries to utilize unlabeled data by predicting their labels. Some studies \citep{zhou2010semi,qiao2018deep} train multiple learners and exploit disagreements among different learners. Other studies utilize self-training to generate pseudo labels for unlabeled data \citep{zhai2019s4l,noisystudent,chen2020big}. 

Consistency regularization \citep{ladder,mean_teacher,uda,rdrop,he2022galaxy} is another popular scheme.
It regularizes the model to be invariant to small perturbations on the input, hidden states, or model parameters.
Mean Teacher \citep{mean_teacher} is an efficient method to implement consistency regularization, where a student model and a teacher model are maintained to enforce the predictions' consistency.
This paper combines pseudo labeling with a teacher-student framework to handle semi-supervised learning in the lifelong scenarios.

\begin{figure}[t]
    \centering
    \includegraphics[width=0.40\textwidth]{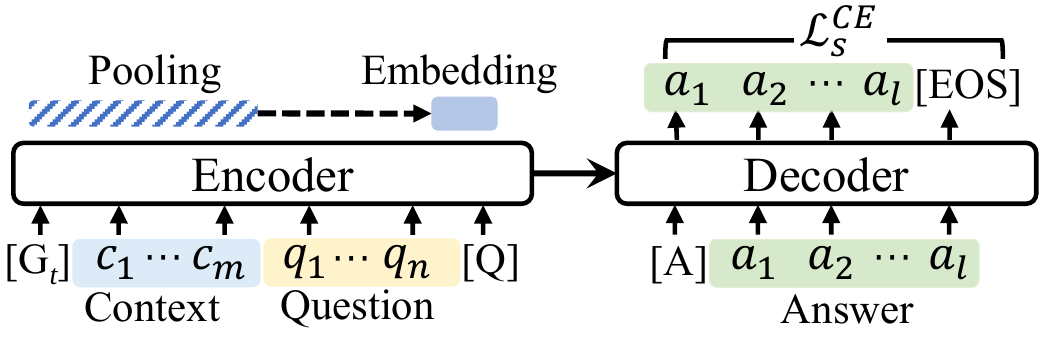}
    \caption{The format of UELL's input-output. ``[G$_t$]'' is the task-specific generation token for task $\mathcal{T}_t$, ``[Q], [A]'' are special tokens that indicate the end of the question and the beginning of the answer, respectively.}
    \label{fig:enc_dec}
\end{figure}

\begin{figure*}[t] 
    \centering
    \includegraphics[width=380px]{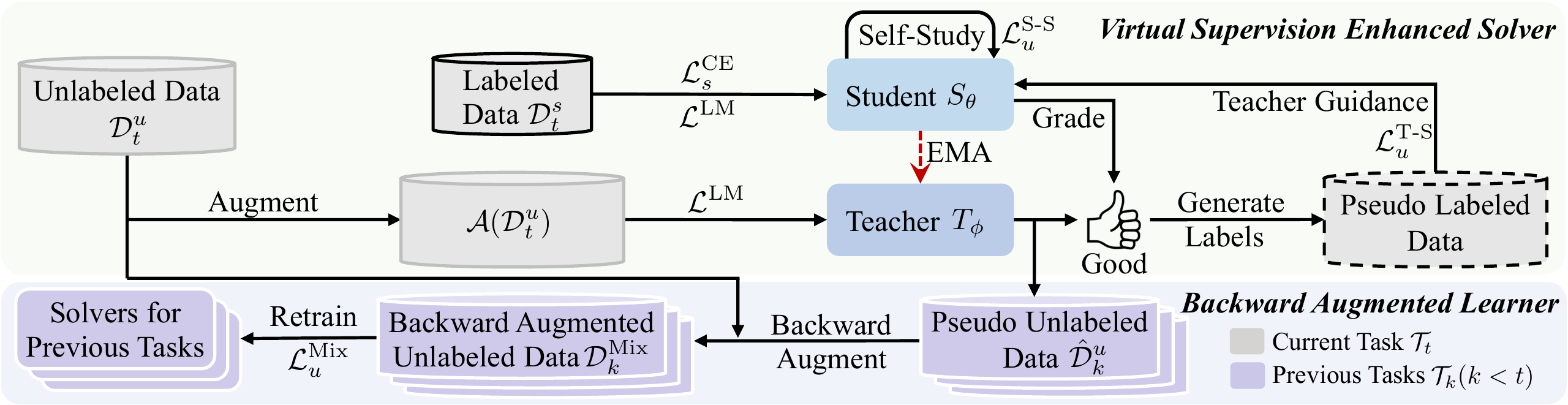}
    \caption{Overview of our method UELL. UELL consists of two modules: a virtual supervision enhanced solver to exploit unlabeled data and a backward augmented learner to encourage knowledge transfer to previous tasks.} 
    \label{fig:whole_model}
\end{figure*}

\paragraph{Lifelong Learning}
aims to learn a sequence of tasks without forgetting previously learned knowledge. 
Three categories of approaches are generally used in lifelong learning:
\textit{Regularization-based} methods either impose constraints on the variation of important weights when learning new tasks \citep{ewc,mas}, or introduce knowledge distillation to preserve the previously learned knowledge \citep{lwf,dhar2019learning};
\textit{Replay-based} methods store real samples \citep{icarl} or generate pseudo samples \citep{lfpt5,zhao2022cvae} for learned tasks to consolidate previous knowledge;
\textit{Architecture-based} methods construct task-specific modules to preserve knowledge.
Some studies \citep{hat,pathnet} use static architectures and apply task-specific routes through the architectures to prevent forgetting, while other studies \citep{tod,adapter_composi,dai2022lifelong} dynamically expand the model with task-specific parameters.

Recently, some studies have tried to investigate semi-supervised lifelong learning for image classification tasks by 
modeling data distributions with generative adversarial networks \citep{gan}, or relying on a super-class structure of image datasets to exploit unlabeled data \citep{disco,hypersemi,memorysemi}.
However, it is non-trivial to extend these works into language tasks since they either require continuous input spaces or utilize extra super-class structures, and all these works do not consider backward knowledge transfer with the help of unlabeled data.

\section{Task Definition and Formulation}
In SSLL, we sequentially learn a stream of semi-supervised language tasks $\mathcal{T}_1$, $\mathcal{T}_2$, $\dots$, $\mathcal{T}_N$.  The data of each task $\mathcal{T}_t$ contain limited labeled data $\mathcal{D}^s_t=\{(X^s_i,Y^s_i)\}$ and abundant unlabeled data $\mathcal{D}^u_t=\{X_i^u\}$, where $X^s_i$ and $X_i^u$ are input samples and $Y^s_i$ is the output label of $X^s_i$. Moreover, the data for task $\mathcal{T}_t$ arrive after task $\mathcal{T}_{t-1}$ is learned \citep{hypersemi}, and we have no access to previous tasks' data. The ﬁnal goal is to optimize the model’s average performance on \textbf{all tasks} after training the whole sequence \citep{lamol}.

Inspired by \citet{decanlp}, we frame different types of NLP tasks into a unified text-to-text format. Specifically, for each $(X^s, Y^s) \in \mathcal{D}^s$, we format $X^s$ as a concatenation of a context and a question, and serialize $Y^s$ as an answer sequence representing the label of $X^s$ (Fig.~\ref{fig:enc_dec}).
For instance, a sample in a sentiment classification task contains the input $X^s$: ``\texttt{I enjoyed the movie.} (context) \texttt{What's the sentiment?} (question)'' and the output $Y^s$: ``\texttt{positive}''.
For each sample $X^u \in \mathcal{D}^u$, the input $X^u$ is only a context.

\section{Methodology}
\subsection{Overview}
We propose an unlabeled data enhanced lifelong learner (UELL) to handle the newly proposed semi-supervised lifelong language learning (SSLL) setting (see Fig.~\ref{fig:data_stream} and~\ref{fig:whole_model}). 
To alleviate the catastrophic forgetting issue of lifelong learning, we dynamically expand the model architecture by allocating task-specific modules.
To overcome the challenges of SSLL discussed in \secref{sec:intro}, we design two modules in UELL:
(1) a virtual supervision enhanced solver exploits unlabeled data for each sequentially arrived task; 
(2) a backward augmented learner encourages knowledge transfer to previous tasks with unlabeled data. 
We will first illustrate the architecture we design to prevent forgetting and then elaborate on details about our learning process. 

\begin{figure}[H]
    \centering
    \includegraphics[width=0.28\textwidth]{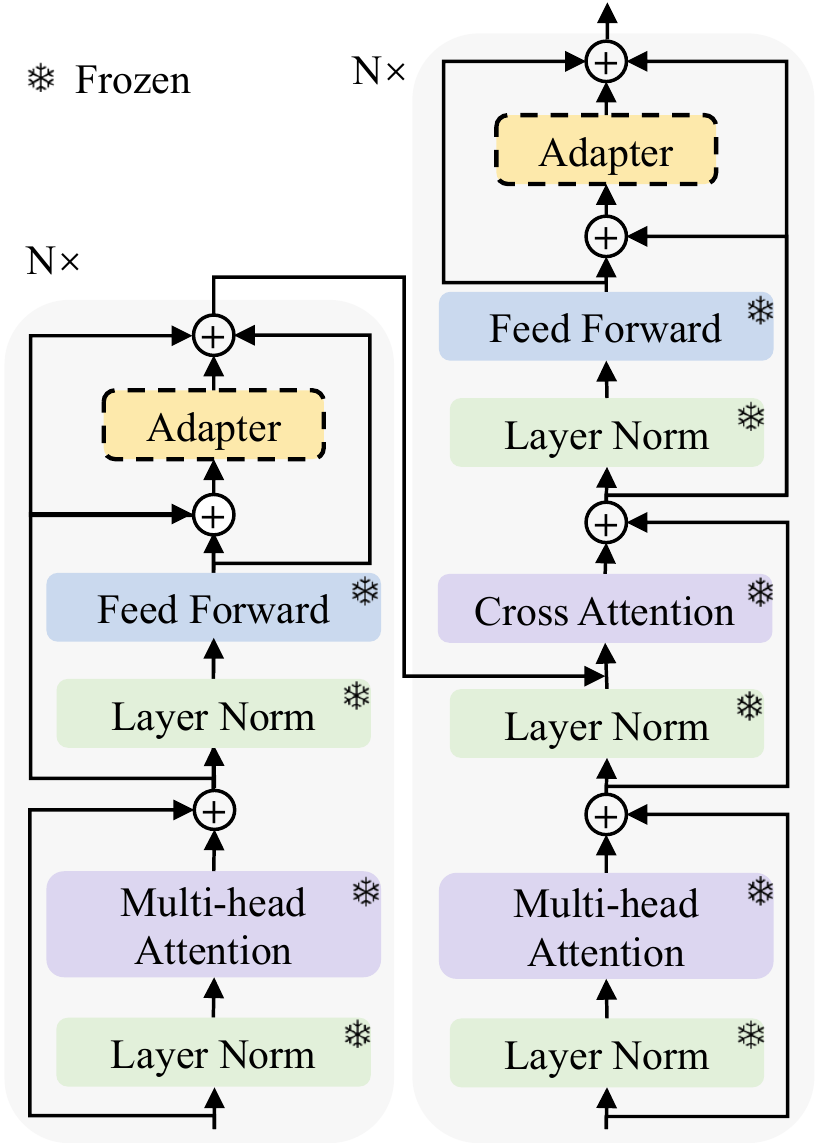}
    \caption{Adapter modules for T5 transformers layers used in UELL. The parameters of the pre-trained T5 (i.e., layers marked with {\small \SnowflakeChevron}) are fixed during training.}
    \label{fig:adapter}
\end{figure}

\paragraph{Model Architecture}
UELL maintains a teacher model $T_{\phi}$ and a student model $S_{\theta}$ for each task during the lifelong learning process. 
It uses T5 \citep{t5} as the backbone for $T_{\phi}$ and $S_{\theta}$ to tackle text-to-text generation tasks.
To overcome catastrophic forgetting, we freeze the parameters of the pre-trained T5 and insert separate adapter modules \citep{adapter,adapterhub} into T5 for each task (See Fig~\ref{fig:adapter}).
Specifically, we use the bottle-necked adapter structure proposed by \citet{pfeiffer2020adapterfusion}, 
in which the adapter consists of a layer normalization \citep{layernorm} followed by a two-layer MLP and a residual connection \citep{he2016deep}.
Such the adapters are light-weighted (i.e., ~0.8\% of T5 parameters). By encapsulating task-specific information into isolated parameters, these adapters can effectively avoid catastrophic forgetting with little memory overhead.

\paragraph{Labeled Data Learning}
We optimize the student model $S_{\theta}$ over $\mathcal{D}^s_t$ for each arriving task $\mathcal{T}_t$ with the following cross-entropy loss (see Fig. \ref{fig:enc_dec}):
\vspace{-0.5mm}
\begin{equation}
\small
\mathcal{L}_s^{\text{CE}} = - \sum_{(X,Y) \in \mathcal{D}^s_t}\log S_\theta(Y|X),
\label{eq:label_CE}
\end{equation}
\vspace{-0.5mm}
The teacher $T_{\phi}$ is gradually updated by means of momentum according to the student's weights $\theta$. We will elaborate on it in \secref{sec:update_procedure}.

\subsection{Virtual Supervision Enhanced Solver}
To fully exploit unlabeled data for each task, a virtual supervision enhanced solver is constructed with a teacher-student framework:
(1) The teacher $T_{\phi}$ predicts pseudo labels for unlabeled data and uses these virtual supervisory signals to guide the student;
(2) The student $S_{\theta}$ takes a self-study course with virtual signals to further enhance itself.

\paragraph{Teacher Guidance} 
\label{para:tea_gui}
Inspired by \citep{mean_teacher}, 
the guidance from the teacher $T_{\phi}$ to the student $S_{\theta}$ is achieved by forcing the consistency between predictions of $T_{\phi}$ and $S_{\theta}$ on perturbed unlabeled inputs.
Specifically, when learning the task $\mathcal{T}_t$, we augment each unlabeled sample $X^u \in \mathcal{D}^u_t$ to $\mathcal{A}(X^u)$ as the input of the teacher, where $\mathcal{A}$ refers to the data augmentation operation that injects noise to the input while preserving its semantics.
Next, the teacher $T_{\phi}$ predicts the pseudo label $\hat{Y}$ of $\mathcal{A}(X^u)$ through greedy decoding.
To alleviate noises introduced by pseudo labels, we only maintain high confident predictions produced by $T_{\phi}$ to guide $S_{\theta}$.
Following \citet{tod}, we use the perplexity score $\text{PPL}(\hat{Y})$ to measure $T_{\phi}$'s confidence when predicting $\hat{Y}$.
Here, low perplexity corresponds to high confidence of $\hat{Y}$.
We optimize $S_{\theta}$ on the filtered pseudo labels using:
\vspace{-0.5mm}
\begin{equation}
\small
 \mathcal{L}_u^{\text{T-S}}\!(\mathcal{D}^u_t)\!=\! - \!\!\!\!\sum _{X^u\in \mathcal{D}^u_t}\!\!\! \!\!\mathbb{1}(\text{PPL}(\hat{Y}) \!\leq \!\tau)
 \cdot\log S_\theta (\hat{Y}|\mathcal{A}(X^u)),
 \label{eq:t-s_loss}
\end{equation}
\vspace{-0.5mm}
where $\tau$ is a confidence threshold.

Moreover, to better guide the student $S_{\theta}$, we argue that the teacher $T_{\phi}$ should adjust its teaching pace based on the learning progress of $S_{\theta}$.
Inspired by \citet{feedback}, we use the training loss of $S_{\theta}$ on labeled data as its grade to inform the teacher.
Considering that unlabeled data are usually ``harder'' to learn than labeled data,
we teach $S_{\theta}$ with unlabeled data only when the student acquires a certain grade (i.e., knowledgeable enough).
Specifically, the optimization of $\mathcal{L}_u^{\text{T-S}}$ is carried out only when the absolute difference of losses between $S_{\theta}$ and $T_{\phi}$ on labeled data drops to a certain threshold $\gamma$.

\paragraph{Student Self-Study}
\label{para:stu_self}
In addition to the teacher's guidance, self-study is also crucial for the student to better leverage unlabeled data.
The self-study course taken by the student is performed by enforcing the student's consistency under small disturbances using a dropout-based regularization term \citep{rdrop}.
Specifically, for each pseudo labeled data ($\mathcal{A}(X^u), \hat{Y}$), we forward $\mathcal{A}(X^u)$ twice through the student model $S_{\theta}$ with different dropout masks to obtain two different predicted outputs:
$S_{\theta}^i(\hat{Y}|\mathcal{A}(X^u))$, ($i=1,2$).
Denote $w_k$ as the $k$-th token in $\hat{Y}$, $S_{\theta}^i(w_k)$ as the predicted distribution for token $w_k$ in the $i$-th forward pass of $S_\theta$.
We optimize the following bidirectional Kullback-Leibler (KL) divergence:
\begin{equation}
\small
\begin{aligned}
\mathcal{L}_u^{\text{S-S}} (\mathcal{D}^u_t) &= \frac{1}{2} \sum_{X^u \in \mathcal{D}^u_t} KL^{12}(X^u) + KL^{21}(X^u), \\
KL^{ij}(X^u) &= \frac{1}{|\hat{Y}|} \sum_{k=1}^{|\hat{Y}|} KL(S_{\theta}^i(w_k) || S_{\theta}^j(w_k)), \\
\end{aligned}
\label{eq:s-s_loss}
\end{equation}
where $KL$ measures the KL divergence between two distributions.

\subsection{Backward Augmented Learner}\label{sec:back_aug}
Besides enhancing the current task with virtual supervision on unlabeled data, 
we find that unlabeled data can be used to encourage backward knowledge transfer.
We build a backward augmented lifelong learner to leverage the newly arrived unlabeled data to improve solvers for previously learned tasks.
This scheme contains two steps: (1) acquiring unlabeled data for learned tasks (\secref{sec:old_unlabel}), and (2) retraining previous solvers (\secref{sec:old_retrain}).

\subsubsection{Previous Unlabeled Data Acquisition}
\label{sec:old_unlabel}
Considering that unlabeled data usually contain rich semantic information, the knowledge of subsequent tasks may facilitate previously learned tasks.
Specifically, when the unlabeled data $\mathcal{D}^u_k$ of the previous task $\mathcal{T}_k~(k<t)$ share similar distributions with the current task $\mathcal{T}_t$, 
we can augment $\mathcal{D}^u_k$ by retrieving similar samples from the unlabeled data $\mathcal{D}^u_t$ of the currently arrived task $\mathcal{T}_t$ based on samples in $\mathcal{D}^u_k$.
However, it is non-trivial to implement the above augmentation process in SSLL because $\mathcal{D}^u_k~(k<t)$ are unavailable when learning $\mathcal{T}_t$ (we have no access to previous data). 

To tackle the above issue, we equip UELL with the ability to generate pseudo unlabeled data that obey the distribution of $\mathcal{D}_k^u$.
In this way, we can retrieve $\mathcal{D}^u_t$ using pseudo unlabeled samples of previous tasks to achieve the aforementioned backward augmentation.
This data generation process is optimized through the language modeling loss on both labeled and unlabeled data as follows,
\begin{equation}
    \small
   \mathcal{L}^{\text{LM}}\!=\!-\!\!\sum_{X\in \mathcal{D}^s_i}\!\! \log S_\theta(X|G_t) \!-\!\mu \!\!\sum _{X\in \mathcal{D}^u_i} \!\!\log S_\theta(X|G_t),
   \label{eq:lm_loss}
\end{equation}
where $G_t$ is a task-specific generation token for task $\mathcal{T}_t$, and $\mu$ is the weight for the unlabeled loss.
$\mathcal{L}^{\text{LM}}$ is only optimized on the context tokens of $X$. 
Then, by feeding the generation token $G_k$ to the encoder of T5, we can sample pseudo unlabeled data of the previous task $\mathcal{T}_k$ from the decoder with the top-K sampling scheme. 

Note that UELL does not further predict labels for the generated unlabeled data to avoid accumulating errors introduced by noisy pseudo-labels.
Moreover, the generated pseudo data are not aimed to prevent forgetting because task-specific adapter modules in UELL are efficient enough to avoid interference among tasks.

\subsubsection{Previous Solvers Retraining}
\label{sec:old_retrain}
When learning the current task $\mathcal{T}_t$, UELL first generates a set of pseudo samples $\{\hat{\mathcal{D}}^u_k\}_{k=1}^{t-1}$ for all previously learned tasks $\{\mathcal{T}_k\}_{k=1}^{t-1}$.
To achieve the backward augmentation for each learned task, we utilize samples in $\hat{\mathcal{D}}^u_k$ to retrieve semantically similar unlabeled data in $\mathcal{D}^u_t$ of task $\mathcal{T}_t$.
Specifically, UELL encodes the sample contexts into representations through the T5 encoder 
\footnote{Here, we use the fixed pre-trained T5 encoder without adapters here to prevent the representations from drifting as new tasks are learned.}
with an average pooling layer (see Fig.~\ref{fig:enc_dec}). Cosine similarities of these representations are used to measure the distance of samples.
For each sample $X^u \in \hat{\mathcal{D}}^u_k$, we retrieve $K$ nearest neighbors from $\mathcal{D}^u_k$ and augment $\hat{\mathcal{D}}^u_k$ with the $K \cdot|\hat{\mathcal{D}}^u_k|$ retrieved neighbors to produce a set of backward augmented unlabeled data $\mathcal{D}^{\text{Mix}}_k$.
Then $\mathcal{D}^{\text{Mix}}_k$ is used to enhance the learned solver for $\mathcal{T}_k$ by optimizing the losses on Eq. \ref{eq:t-s_loss} and \ref{eq:s-s_loss}:
\begin{equation}
\vspace{-5pt}
\small
    \mathcal{L}_u^{\text{Mix}}(\mathcal{D}^\text{Mix}_k)=\mathcal{L}_u^{\text{T-S}}(\mathcal{D}_k^\text{Mix})+\mathcal{L}_u^{\text{S-S}}(\mathcal{D}_k^\text{Mix}).
    \vspace{-5pt}
    \label{eq:mix_loss}
\end{equation}
In this way, we can encourage the knowledge transfer from newly encountered unlabeled data to previously learned tasks.

\begin{algorithm}[t]
  \algsetup{linenosize=\tiny} 
  \small
  \caption{UELL Training}
  \label{al:train}
  \begin{algorithmic}[1]
    \STATE \textbf{Input:}
    Semi-supervised tasks $\{\mathcal{T}_t\}_{t=1}^N$, 
    A pretrained T5 model, and randomly initialized student and teacher with parameters $\theta_0$ and $\phi_0$, respectively. 
    A learning rate $\eta$, EMA decay rate $\alpha$.
    \STATE \textbf{Output:} Learned teacher parameters $\theta_t$ for tasks $\{\mathcal{T}_t\}_{t=1}^N$
    \FOR{$t=1$ to $N$} 
    \STATE Initialize student $\phi_t$ and teacher $\theta_t$ using $\phi_{t-1}$
    \WHILE{Not Converge}
    \STATE Sample batches $B^s \subseteq \mathcal{D}^s_t$ and $B^u \subseteq \mathcal{D}^u_t$.
    \STATE Compute $\mathcal{L}_s^{\text{CE}}$ (Eq.\ref{eq:label_CE}) and $\mathcal{L}^{\text{LM}}$ (Eq.\ref{eq:lm_loss}).
    \IF{student $\theta_t$ reaches grade $\gamma$}
    \STATE Augment $B^u$ to $\mathcal{A}(B^u)$.
    \STATE Predict $\hat{Y}$ for each $X \in \mathcal{A}(B^u)$ using $T_{\phi_t}$.
    \STATE Compute $\mathcal{L}_u^{\text{T-S}}$ (Eq.\ref{eq:t-s_loss}) and $\mathcal{L}_u^{\text{S-S}}$ (Eq.\ref{eq:s-s_loss}).
    \ENDIF
    \STATE Compute the total loss $\mathcal{L}$ (Eq.~\ref{eq:total_loss}).
    \STATE Update student $\theta_{t}\leftarrow \theta_{t}-\eta \nabla\mathcal{L}$.
    \STATE Update teacher $\phi_{t}\leftarrow \alpha \phi_{t}+(1-\alpha)\theta_{t}$.
    \ENDWHILE
    \FOR{$k=1$ to $t-1$}
    \STATE Generate pseudo data $\hat{\mathcal{D}}_k^u$ for task $\mathcal{T}_k$.
    \STATE Backward augment $\hat{\mathcal{D}}_k^u$ to $\mathcal{D}_k^{\text{Mix}}$
    \STATE Update student $\theta_{k}$ and teacher $\phi_{k}$ with $\mathcal{L}_u^{\text{Mix}}$ (Eq. \ref{eq:mix_loss})
    \ENDFOR
    \ENDFOR
  \end{algorithmic}
  \end{algorithm}

\subsection{Model Update Procedure}
\label{sec:update_procedure}
Before learning the first task, the teacher and student models in UELL are initialized with randomly-initialized adapters layers with the pre-trained T5 backbone.
When learning the current task $\mathcal{T}_t$, the student model $S_\theta$ is trained using the following loss:
\begin{equation}
\small
\mathcal{L}=\mathcal{L}_s^{\text{CE}} + \mu \mathcal{L}_u^{\text{T-S}}(\mathcal{D}_t^u) + \mu \mathcal{L}_u^{\text{S-S}}(\mathcal{D}_t^u) +\lambda \mathcal{L}^{\text{LM}},
\label{eq:total_loss}
\end{equation}
where $\lambda$ is the weight to balance the task learning and language modeling.
To prevent confirmation bias \cite{mean_teacher}, the teacher weights $\phi$ are updated as an exponential moving average (EMA) of student weights in each batch:
\begin{equation}
\phi_{p}=\alpha \phi_{p-1}+(1-\alpha)\theta_{p}, \label{eq:ema}
\vspace{-5pt}
\end{equation}
where $\alpha$ is the EMA decay rate, $p$ is the time step. Note that the slowly evolved teacher can be regarded as an ensemble of student models in different training iterations. This leads to more stable and accurate predictions on unlabeled data \citep{mean_teacher}.
After learning task $\mathcal{T}_t$, UELL generates pseudo unlabeled data for each previously learned tasks $\{\mathcal{T}_k\}_{k=1}^{t-1}$ and further optimizes the learned solvers with the loss $\mathcal{L}^{\text{Mix}}_u(\mathcal{D}^{\text{Mix}}_k)$ shown in Eq.\ref{eq:mix_loss} to enable the backward knowledge transfer.
See Algorithm~\ref{al:train} for more details.

\section{Experiment Setup}
\subsection{Datasets}
Following \citet{lamol}, we evaluate our approach from two dimensions: (1) tasks with the Same Type but Different Domains (STDD); (2) tasks of Different Types (DT).
For STDD, we follow \citet{lamol} to use five text classification datasets covering domains from news classification, sentiment analysis, and Wikipedia article classification. We follow \citet{episodic} to produce balanced datasets.
For DT, we consider five different sequence generation tasks from decaNLP \citep{decanlp}: question answering, semantic parsing, semantic role labeling, goal-oriented dialogue generation, and sentiment analysis.
To each task $\mathcal{T}_t$ in the SSLL setting, we randomly select 100 labeled data to construct $\mathcal{D}^s_t$ and select another 2,000 unlabeled data to construct $\mathcal{D}^u_t$.
More details are provided in Table~\ref{tab:datasets} and Appendix~\ref{appn:dataset_metric}.

\begin{table*}[t]
\centering
\small
\begin{tabular}{cllcccc}
\toprule
Dimensions & Datasets & Metrics  & \# Training Set & \# Testing Set \\
\midrule
\multirow{5}{*}{STDD} & AGNews & EM  & 115,000 & 7,600  \\
                      & Amazon & EM  & 115,000 & 7,600  \\
                      & DBPedia & EM & 115,000 & 7,600   \\
                      & Yahoo & EM   & 115,000 & 7,600  \\
                      & Yelp & EM    & 115,000 & 7,600  \\
\midrule
\multirow{5}{*}{DT} & SQuAD & nF1     & 87,599 & 10,570   \\
                    & WikiSQL & lfEM  & 56,355 & 15,878     \\
                    & SST     & EM   & 6,920   & 1,821    \\
                    & QA-SRL  & nF1  & 6,414   & 2,201    \\
                    & WOZ     & dsEM  & 2,536  & 1,646        \\
\bottomrule
\end{tabular}
\caption{Summary of datasets statics and their metrics. nF1 is the normalized version of the F1 score; EM represents an exact match between texts: for text classification tasks) this amounts to accuracy; for WOZ, it is equivalent to dfEM (turn-based dialogue state exact match); for WikiSQL, it is equivalent to lfEM (exact match of logical forms).}
\label{tab:datasets}
\end{table*}

\subsection{Implementation Details}
\label{sec:imple}
We use T5-base \citep{t5} as our backbone and implement adapters using AdapterHub \citep{adapterhub}. 
We set the confidence threshold $\tau=1.5$ (Eq.\ref{eq:t-s_loss}), unlabeled loss weight $\mu=0.01$ (Eq.\ref{eq:lm_loss}), language modeling loss weight $\lambda=0.5$ (Eq.\ref{eq:total_loss}), and EMA decay rate $\alpha=0.95$ (Eq.\ref{eq:ema}).
We set the threshold of teacher guidance $\gamma$ to $0.1$ in \secref{para:tea_gui} and choose $K=3$ nearest neighbors in \secref{sec:back_aug}.
We train our model UELL on 1 Tesla-V100 GPU. 
Each task in STDD and DT is trained using the Adam optimizer \citep{adam} for 120 and 200 epochs, respectively, with a warm-up ratio of 0.1 and maximum learning rate of 2e-4.
It takes around 5 and 18 hours to learn all STDD tasks and DT tasks, respectively.
The training and testing batch size is set to 16. We use EDA \citep{eda} to implement the data augmentation on unlabeled data $\mathcal{D}^u$ as $\mathcal{A}(\mathcal{D}^u)$.
For backward augmentation, we train one epoch on the augmented unlabeled data $\mathcal{D}^{\text{Mix}}_k$ for previous task $\mathcal{T}_k$ to optimize the previously learned solver.

All results reported in this paper are averages of five different runs with random task orders.

\subsection{Baselines}
We compare our model with the following baselines.
\textbf{Fine-tuning} (FT) directly tunes a pretrained T5 model on incoming data.
\textit{Regularization-based methods}: \textbf{EWC} \citep{ewc} and \textbf{MAS} \citep{mas} mitigate forgetting by penalizing variation of important parameters for previous tasks;  
\textit{Replay-based methods}: \textbf{ER} \citep{er} stores real samples of learned tasks to prevent forgetting.
\textbf{LAMOL}\citep{lamol} generates pseudo samples of previous tasks and trains them with the new tasks' data;
\textit{Architecture-based methods}:
\textbf{HAT} \citep{hat} uses a task-specific hard attention mechanism to preserve previously learned tasks.
\textbf{CTR} introduces continual learning plugins into BERT and uses task masks to preserve task-specific knowledge and encourage knowledge transfer \footnote{CTR and HAT are specifically designed for classification tasks, and it is non-trivial to extend them to text-generation tasks, so we do not compare them for DT tasks.};
\textbf{Adapter} \citep{tod} dynamically expands the model by assigning task-specific adapters. For fair comparisons, we use a pretrained T5 as its backbone.
\textbf{Compositional-Adapters} (Comp) \citep{adapter_composi} utilizes hidden state mixing to adaptively compose old and new adapters for new tasks and employs generative replay to facilitate knowledge transfer.
Besides the above baselines, we also test the \textbf{Multi-task Learning} (MTL) approach that tunes the whole model to learn all tasks simultaneously. This approach is usually seen as an upper bound of lifelong learning.

Note that all the above baselines focus on supervised lifelong learning.
To enable a fair comparison under the SSLL setting, we enhance the baselines with a strong pseudo labeling method to utilize unlabeled data \citep{noisystudent}.
Specifically, for each task, we first train each baseline model on labeled data. We generate pseudo labels for unlabeled data and train the mixture of labeled data and pseudo labeled data as the final model.
For compared baselines, we follow their original settings for training.
See more details in Appendix~\ref{appn:imple}.

\subsection{Metrics}
Following \citet{lamol}, we evaluate each task with its corresponding metric (see Table~\ref{tab:datasets}).
The score of each metric lies from 0 to 100\%. 
We evaluate the performance of lifelong learning using \textit{Average Score} (\textbf{Avg-Score}) \citep{lamol,tod}
that measures the average test scores of all learned $N$ tasks: 
\vspace{-10pt}
\begin{equation*}
\small
\text{Avg-Score}=\frac{1}{N}\sum _{j=1}^{N}R_{N,j},
\vspace{-8pt}
\end{equation*}
where $R_{i,j}$ is the test score of task $\mathcal{T}_j$ after the $i$-th task is learned.
Following \citet{gem}, we evaluate the effect of \textit{backward knowledge transfer} (\textbf{BWT}) to assess the impact of learning on subsequent tasks on previously learned tasks: 
\vspace{-10pt}
\begin{equation*}
\small
\text{BWT}=\frac{1}{N-1}\sum_{j=1}^{N-1}R_{N,j}-R_{j,j}.
\vspace{-8pt}
\end{equation*}
A negative BWT indicates that the model has forgotten some previously acquired knowledge, i.e., suffers from catastrophic forgetting.
In general, the higher of these two metrics, the better the model.

\begin{table*}[t]
    \centering
    \resizebox{0.92\textwidth}{!}{
    \begin{tabular}{lcccccccccc>{\columncolor{lightgray}}c}
    \toprule
    & FT & EWC & MAS & LAMOL & ER & HAT & CTR & Adapter & Comp & \textbf{UELL} & MTL \\
    \midrule
    Avg T.P. &  223M & 125M & 125M & 125M & 223M & 73M & 75M & 1.79M & 2.44M & 1.79M & 223M \\
    \midrule
     Avg-Score  &  21.34 & 45.85 & 48.01 & 48.21 & 64.47 & 55.24 & 52.07 & 69.05 & 48.13 & \textbf{71.12} & 75.67  \\
     BWT        &  -58.24 & -13.31 & -13.26 & -6.611 & -4.012 & -3.727 & -2.310 & 0.000 & -18.42 & \textbf{0.146} & N/A  \\
     \bottomrule 
    \end{tabular}}
    \caption{Results of five SSTD tasks. Each result is an average of five random task orders. 
UELL is significantly better than other SSLL baselines with $p$-value $\leq 0.05$ under $t$-test. ``Avg T.P.'' refers to the number of tunable parameters for each task.}
    \label{tab:main_stdd}
\end{table*}

\begin{table*}[t]
\centering
\resizebox{0.78\textwidth}{!}{
\begin{tabular}{ccccccccc>{\columncolor{lightgray}}c}
\toprule
            & FT & EWC & MAS & LAMOL & ER & Adapter & Comp & \textbf{UELL} & MTL \\
\midrule
  Avg-Score & 25.14 & 30.31 & 34.88 & 48.60 & 50.91 & 65.99 & 43.46 & \textbf{70.75} & 75.83  \\
  BWT       & -52.14 & -28.62 & -22.13 & -2.590 & -13.92 & 0.000  & -3.59 & \textbf{0.393} & N/A \\ 
\bottomrule
\end{tabular}}
\caption{Results of five DT tasks. Each result is an average of five random task orders. UELL is significantly better than other SSLL baselines with $p$-value $\leq 0.05$ under $t$-test.}
\label{tab:main_dt}
\end{table*}

\section{Experimental Results and Analyses}
\subsection{Main Results}
Table~\ref{tab:main_stdd} and~\ref{tab:main_dt} show the performances of all methods.
UELL significantly outperforms all baselines by a large margin. 
We can also observe that: 
\begin{itemize}[leftmargin=*]
\setlength{\itemsep}{0pt}
\setlength{\parsep}{0pt}
\setlength{\parskip}{0pt}
\item Directly tuning a single model sequentially (the FT baseline) suffers from severe catastrophic forgetting issues, highlighting the importance of lifelong learning studies.
\item Regularization based methods EWC and MAS improve the lifelong learning performance to some extent, but they still perform inferior to replay-based methods ER and LAMOL. 
\item ER outperforms LAMOL on the Avg-Score, indicating that real samples carry higher-quality knowledge than pseudo samples. This validates our approach of using real samples obtained for the current task to transfer knowledge backward to previous tasks.
\item Tuning one model for all tasks with methods like ER, LAMOL and HAT still faces the issue of forgetting previously learned knowledge (i.e., negative BWT). 
\item CTR and Comp bring negative interference among tasks, resulting in poor performances. 
This validates the effectiveness of assigning task-specific parameters for lifelong learning tasks (Adapter and UELL).
\item The higher performance of UELL compared to Adapter indicates that our method makes better use of unlabeled data, and the introduced backward augmentation does transfer knowledge from newly arrived tasks to learned tasks.
\end{itemize}

\subsection{Ablation Studies}
We conduct ablation studies to verify the effectiveness of each proposed component in UELL. The task orders of STDD and DT are randomly selected.
(1) \textbf{w/o Unlabel} means no unlabeled data is utilized for each task. 
Here, we investigate whether unlabeled data bring benefits to supervised lifelong learning.
(2) \textbf{w/o Selection} skips the confidence selection process and uses all predicted pseudo labeled data to teach the student, i.e., set $\tau=\infty$ in Eq.\ref{eq:t-s_loss};
(3) \textbf{w/o Interact} ignores the interaction between the teacher and student, i.e., the loss $\mathcal{L}_u^{\text{T-S}}$ is optimized in every training iteration. 
(4) \textbf{w/o Self-Study} removes the self-study loss $\mathcal{L}_u^{\text{S-S}}$ (Eq.\ref{eq:s-s_loss}) for the solver optimization;
(5) \textbf{w/o Back-Aug} means that newly arrived unlabeled data are not utilized to promote the learned solvers, i.e., schemes introduced in \secref{sec:back_aug} are ignored.

From Table~\ref{tab:abaltion}, we can see that: (1) Unlabeled data leveraged by UELL greatly improve the supervised lifelong learning. This validates our claim that UELL can effectively overcome the first challenge of SSLL, i.e., exploiting unlabeled data.
(2) Removing the pseudo-label selection process makes our student suffer from noisy pseudo labels predicted by the teacher, thus downgrading their lifelong learning performance. This also validates the effectiveness of our confidence-based label selection scheme.
(3) Removing interaction between the teacher and student degenerates the performance. It verifies that learning harder and noisier knowledge from unlabeled data is non-trivial.
(4) Self-study of the student indeed enhances its capability of utilizing unlabeled data.
(5) The backward augmentation encourages the knowledge transfer from new tasks to old tasks, further promoting overall performance.
This validates our claim that UELL can effectively tackle the second challenge of SSLL, 
i.e., leverage unlabeled data to encourage knowledge transfer to previous tasks.

\begin{table*}[t]
\small
\centering
\resizebox{0.92\textwidth}{!}{
\begin{tabular}{llcccccc}
\toprule
& & \textbf{UELL} & w/o Unlabel & w/o Select & w/o Interact & w/o Self-Study & w/o Back-Aug \\
\midrule
\multirow{2}{*}{STDD}    & Avg-Score & \textbf{71.57} & 68.13 & 70.64 & 68.93 & 70.97 & 69.73    \\
                         & BWT       & \textbf{0.144} & 0.000 & 0.039 & 0.089 & 0.049 & 0.000 \\ 
\midrule
\multirow{2}{*}{DT} & Avg-Score & \textbf{73.10} & 67.34 & 71.47 & 71.82 & 71.42 & 72.80   \\
                    & BWT       & \textbf{0.713} & 0.000 & 0.189 & 0.021 & 0.024 & 0.000 \\ 
\bottomrule
\end{tabular}}
\caption{Ablation studies on STDD and DT tasks. ``w/o'' means removing the corresponding component in UELL.}
\label{tab:abaltion}
\end{table*}
    
\subsection{Data Efficiency Analyses}
We analyze the data efficiency of UELL by fixing the unlabeled data and vary the number of labeled data on SSLL.
We randomly select a task order from STDD to conduct the analyses.
As shown in Table~\ref{tab:efficiency}, the Avg-Score of UELL increases with the amount of labeled data.
The performance of UELL does not significantly degenerate in the few-shot setting.
Even with only 50 labeled samples, UELL can still surpass all the baselines that use 100 labeled data in Table \ref{tab:main_stdd}.
Moreover, we also notice that the BWT score of UELL trained with 2000 labeled samples is lower than that of 100 labeled data.
We speculate this is because our solver has acquired sufficient knowledge from labeled data, and the knowledge it can gain from unlabeled data of subsequent tasks is limited.
In this case, it is hard to perform backward transfer because the learned solvers are already knowledgeable.

Further, we assess the ability of UELL to leverage unlabeled data by varying the amount of unlabeled data while keeping the labeled data fixed. 
As shown in Table~\ref{tab:efficiency}, we can see that UELL gets better performances (i.e., higher Avg-Score and BWT) with more unlabeled data. 
This validates the effectiveness of UELL for leveraging unlabeled data to improve the overall performance of SSLL and encouraging more knowledge transfer from new tasks to previous tasks.
\begin{table}[H]
    \centering
    \resizebox{0.48\textwidth}{!}{
    \begin{tabular}{lccc|cccc}
    \toprule
    & \multicolumn{3}{c}{\# Labeled Data} &  \multicolumn{4}{c}{\# Unlabeled Data}\\
    \cline{2-8}
     & 50 & 100 & 2000    & 0 &  500 & 2000 & 10000 \\
     \midrule
      Avg-Score  & 68.64 & 71.57 & 75.06 & 68.13 & 70.51 & 71.57 & 71.92 \\
        BWT      & 0.056 & 0.144 & 0.016 & 0.000 & 0.059 & 0.144 & 0.223 \\
    \bottomrule
    \end{tabular}}
    \caption{The performances of UELL with different number of labeled and unlabeled data.} 
    \label{tab:efficiency}
\end{table}

\subsection{Longer Task Sequences Analysis}
To verify the ability of UELL to handle more tasks, we combine STDD and DT tasks to form a longer sequence of ten tasks. We compare UELL with its upper bound MTL and three best-performing baselines LAMOL, ER, and Adapter.
We randomly select three task orders and report their average performances in Table~\ref{tab:long}.
Our method UELL still outperforms these baselines with a large margin, suggesting that UELL can be generalized to longer task sequences. 

\begin{table}[H]
    \centering
    \resizebox{0.45\textwidth}{!}{
    \begin{tabular}{lcccc>{\columncolor{lightgray}}c}
    \toprule
    & LAMOL & ER & Adapter &  \textbf{UELL} & MTL \\
    \midrule
       Avg-Score  &  52.71 & 61.68  & 67.52   & \textbf{69.23} & 71.82 \\
       BWT        &  -4.610 & -4.295 & 0.000  & \textbf{0.235} & N/A  \\    
       \bottomrule
    \end{tabular}}
    \caption{Performances of UELL and some baselines under longer task sequence.}
    \label{tab:long}
\end{table}

\subsection{Analyses of Backward Augmentation}
For backward augmentation in \secref{sec:back_aug}, 
the number of neighbors $K$ when retrieving similar data is important.
Hence, we conduct analyses to investigate how the value of $K$ affects UELL's performances. 
A random task order of STDD is selected to implement the analyses.
As shown in Table~\ref{tab:memory_analyses}, the overall performance of UELL fluctuates with the value of $K$.
The backward transfer performance generally improves as $K$ increases. However, if $K$ is too large, we are more likely to absorb and retrieve dissimilar data to $\mathcal{D}_k^{\text{Mix}}$ and thus degenerate the model performance.
\begin{table}[H]
    \centering
    \resizebox{0.32\textwidth}{!}{
    \begin{tabular}{lcccc}
    \toprule
    $K$        & 1 & 3 & 30 & 50  \\
    \midrule
      Avg-Score   & 71.35 & 71.57 & 71.31 & 71.28  \\
        BWT       & 0.092 & 0.144 & 0.158 & 0.123 \\
    \bottomrule
    \end{tabular}}
    \caption{Impacts of the number of neighbors $K$.}
    \label{tab:memory_analyses}
\end{table}

\subsection{Computation Resource Analysis}
We report the number of tunable parameters for UELL and baselines to assess their computation cost (see Tabel \ref{tab:main_stdd}). 
UELL utilizes the smallest tunable parameters but achieves the best performances of SSLL.

\subsection{Case Study of Pseudo Unlabeled Samples}
We present some pseudo unlabeled samples generated by UELL in Appendix~\ref{appn:pseudo}. We can observe that UELL generates high-quality pseudo samples for learned tasks. This benefits from the sufficient data ($\mathcal{D}^s$ and $\mathcal{D}^u$) to learn the language modeling ability for UELL by optimizing $\mathcal{L}^{\text{LM}}$ in Eq.~\ref{eq:lm_loss}.

\section{Conclusion}
In this paper, we propose a new setting, semi-supervised lifelong language learning (SSLL), where a model learns a sequence of language tasks using both labeled and unlabeled data. We build a novel method UELL to tackle challenges in SSLL.
UELL contains a virtual supervision enhanced solver to exploit unlabeled data for each task and a backward augmented learner to encourage knowledge transfer from subsequent tasks to previously learned tasks. Extensive experiments and analyses on language tasks demonstrate the effectiveness of UELL in leveraging unlabeled data, mitigating catastrophic forgetting, and encouraging backward knowledge transfer in the SSLL setting.  

\section*{Limitations}
As our first attempt in the new semi-supervised lifelong language learning (SSLL) setting, our method UELL assumes the unlabeled data of each task are intrinsically related to labeled data.
We have not investigated the unlabeled data from general corpus such as Common Crawl \footnote{Common Crawl link: \url{https://commoncrawl.org}}, BooksCorpus \citep{zhu2015aligning} and Wikipedia \footnote{Wikipedia link: \url{https://huggingface.co/datasets/wikipedia}} to improve lifelong language learning with limited labeled data.

Fortunately, pre-training schemes may already provide insights for the above problems. The pretrained T5 checkpoints we use to initialize the UELL model have been pretrained on these general corpus with well-designed losses. We can explore including these pretraining losses in further attempts for the SSLL setting. 
Pre-training models that are obtained from other corpora \cite{he2022galaxy,zhou2021eva,wang2020large,zheng2020pre,zheng-etal-2022-mmchat,he2022unified,zheng2019personalized} may also help to alleviate this issue.

Moreover, our UELL model constructs task-specific adapter modules to prevent forgetting.
As a similar approach to adapters, the prompt learning also enables us to share the same backbone language model while dynamically allocating task-specific parameters \citep{promptsurvey}. Prompts can also be used to implement parameter-efficient lifelong learning schemes. The virtual supervision enhanced solver and backward augmented learner proposed in UELL can be directly combined with prompt learning based approaches. 
In future works, we aim to explore the prompt-based approach to tackle challenges in SSLL.
Some approaches on few-shot learning \cite{zhao2022improving} can also be applied in our SSLL setting.

\section*{Ethics Statement}
This work does not present any direct ethical issues.
In our paper, the method UELL is proposed to cope with a more realistic setting, semi-supervised lifelong language learning, where the model learns sequentially arrived language tasks that are partially labeled.
All our experiments are conducted on publicly available datasets. 
All terms for using these datasets are strictly followed in our study.
The metrics used in our paper are automatic and do not involve manual labor.

\section*{Acknowledgement}
Research on this paper was supported by Alibaba Group through Alibaba Research Intern Program and Hong Kong Research Grants Council (Grant No. 16204920).

\bibliography{anthology,custom}
\bibliographystyle{acl_natbib}

\clearpage
\appendix
\section{Experiments Details}
\subsection{Datasets and Metrics}
\label{appn:dataset_metric}
Details of the datasets we use in our studies are listed below.
Five datasets (tasks) from decaNLP \citep{decanlp,lamol}:
\begin{itemize}[leftmargin=*]
\item Question Answering – Stanford Question Answering Dataset (SQuAD) \citep{squad}: It consists of context, questions, and answers. The context is paragraphs from English Wikipedia, and the answers are spanned from its corresponding question paragraphs.
\item Semantic Parsing – WikiSQL \citep{wikisql}: WikiSQL provides logical forms along with natural language utterances. 
\item Sentiment Analysis – Stanford Sentiment Treebank (SST) \citep{sst}: It consists of movie reviews with its answers, including positive and negative binary options. 
\item Semantic Role Labeling – QA-SRL \citep{srl}: It is a question answering form of the SRL task. 
\item Goal-Oriented Dialogue – English Wizard of Oz (WOZ) \citep{woz}: WOZ is a restaurant reservation task that provides a predeﬁned ontology of a series of information for helping an agent to make reservations for customers. 
\end{itemize}
Five text classiﬁcation datasets (tasks) from MBPA++ \citep{episodic,lamol}:
\begin{itemize}
\item AGNews: News articles to be classiﬁed into 4 classes.
\item Yelp and Amazon: Customer reviews and ratings on Yelp and Amazon. Both datasets include 5 classes.
\item DBPedia: Articles and their corresponding categories on Wikipedia, including 14 classes.
\item Yahoo: Questions and answers on the Yahoo! platform, including 10 classes.
\end{itemize}

\subsection{Details of Baselines Implementation}
\label{appn:imple}
For LAMOL, we also use task-specific generation tokens to perform the generative replay. The sampling ratio of pseudo samples is set to 0.2.
For ER, we store real samples of labeled data and unlabeled data with a ratio of 0.1 for each task.
To adapt compared baselines to SSLL, models trained on labeled data are used to generate pseudo labels of unlabeled data and are further trained with the mixture of labeled data and pseudo labeled data like \citep{noisystudent}. 

\section{Case Study of Pseudo Unlabeled Data}
\label{appn:pseudo}
We show a few pseudo unlabeled data of three tasks generated by UELL in Table~\ref{tab:pseudo_unlabel}.

\begin{table*}[t]
    \centering
    \begin{tabular}{p{0.8\linewidth}}
    \toprule
    Sentiment Analysis\\
    \midrule
    1. The performance is priceless.       \\
    2. This humbling film, in all its minutiae and in its ambiguity, is simply a waste of money.      \\
    3. This is one of those films that sneaks up on you and stays with you long after you' ve left the theatre.      \\
    \midrule
    News Classiﬁcation \\
   \midrule 
   1. Oil Spite Changes (News) Petroleum companies are changing their share price target to become cheaper, a new research study said Wednesday. \\
   2. In a bid to reach the future of health care, the government recently endorsed the use of the halo method to treat chronic diseases. \\
   3. At least two teams make headlines to help a new scottish league team to win their first title defending champion Al Hamdat this December, as a defensive midfielder. \\
   \midrule
   Goal-Oriented Dialogue \\
   \midrule
   1. I would like a cheap restaurant in the north part of town. \\
   2. Can you tell me the phone number of the Chinese restaurant? \\
   3. Yes, I need the address and phone number please. \\
    \bottomrule
    \end{tabular}
    \caption{Generated pseudo unlabeled data for three tasks.}
    \label{tab:pseudo_unlabel}
\end{table*}
\end{document}